\documentclass[11pt]{article}

\usepackage{acl}

\usepackage{latexsym}
\usepackage{microtype}

\usepackage{fontspec}
\newfontfamily\evfont{LiberationSerif}[
    Path = ./fonts/,
    Extension = .ttf,
    UprightFont = LiberationSerif-Regular,
    BoldFont = LiberationSerif-Bold,
    ItalicFont = LiberationSerif-Italic,
    BoldItalicFont = LiberationSerif-BoldItalic
]
\newcommand{\evchar}[1]{{\evfont #1}}
\usepackage{graphicx}
\usepackage{booktabs}
\usepackage{amsmath}
\usepackage{amssymb}
\usepackage{array}
\usepackage{tabularx}
\usepackage{multirow}
\usepackage[shortlabels]{enumitem}
\usepackage{xcolor}
\usepackage{xspace}
\usepackage{url}
\usepackage{placeins}

\newcommand{\qwen}{Qwen3-VL-4B\xspace}

\title{EvenVLM: A Text and Multimodal Dataset for the Even Language with Vision-Language Fine-Tuning}

\author{Daniil Makovetskii$^{1}$ \and Ivan Bondarenko$^{1}$ \\
        $^{1}$Novosibirsk State University, Novosibirsk, Russia \\
        \texttt{d.makovetskii@g.nsu.ru} \quad \texttt{i.bondarenko@g.nsu.ru}}

\begin{document}
\maketitle

\begin{abstract}
We present EvenVLM, the first text and multimodal (VQA) dataset for Even (Lamut), an endangered Northern Tungusic language spoken by approximately 6,800 people in Siberia.
Our dataset comprises 50,521 structured Even-Russian entries extracted from dictionary and phrasebook sources via GPT-4o Mini-based parsing, from which we synthetically generate 107,543 training examples spanning translation, grammar, reading comprehension, and semantic classification.
To address tokenization inefficiency on Even's agglutinative morphology, we propose a three-step tokenizer expansion algorithm that adds Even-specific graphemes (\evchar{ӈ}, \evchar{ө}, \evchar{ӫ}), whitelisted vocabulary, and BPE merges, reducing average tokens per word by 27\% (4.99 to 3.64).
Using LoRA fine-tuning of Qwen3-VL-4B, we adapt the model for text tasks (0.53\% trainable parameters, rank 32) and separately for visual question answering (rank 64, frozen vision encoder).
Our best configuration achieves 11.4\% exact match and a 13.6 weighted score on text benchmarks, 0.1283 chrF for sentence translation, and 65\% accuracy on VQA (up from 0\% for the base model and text-only fine-tuning).
We release our datasets, evaluation benchmarks, and training pipeline to support further research on Even and other low-resource languages.
\end{abstract}

\section{Introduction}

Even (Lamut) is a Northern Tungusic language spoken by approximately 6,800 people (out of 19,900 who identify as Even) across the Magadan, Kamchatka, and Sakha regions of Siberia, according to the 2020--2021 census~\cite{atlas2025}.
The language is classified as endangered: the number of native speakers has declined sharply over recent decades, and intergenerational transmission is limited.
Despite the growing interest in multilingual AI, very few researchers work on Even in the context of machine learning, natural language processing, or data science.
Consequently, any first steps in this direction---whether dataset creation, model adaptation, or evaluation---are valuable for documenting, preserving, and supporting the Even language.
While large vision-language models (VLMs) have shown remarkable capabilities in high-resource languages~\cite{qwen2025}, their application to languages with fewer than 10,000 speakers remains virtually unexplored~\cite{lupascu2026}.

This work directly addresses the ORACLE workshop's mission of open reasoning across languages and cultures. We contribute an \textbf{open data and evaluation resource} for a critically underrepresented language, enabling culturally grounded multilingual assessment of vision-language models~\cite{lupascu2026}. By providing evaluation tools in Even itself---rather than through translation to Russian---we support trust in multilingual GenAI for indigenous communities, who can assess model outputs in their own language. Our pipeline is designed to be reproducible and adaptable to other low-resource agglutinative languages.

In this work, we address this gap by fine-tuning \qwen, a 4.4B-parameter VLM, for Even language tasks spanning both textual and multimodal domains.
Our contributions are threefold:

\begin{enumerate}[leftmargin=*,nosep]
    \item \textbf{Dataset.} We introduce the first text and VQA dataset for Even, comprising 50,521 dictionary entries synthetically expanded to 107,543 training examples, and a multimodal set of 189 images with 477 bilingual annotations, producing 1,908 VQA examples.
    \item \textbf{Tokenizer.} We propose a three-step algorithm adapting the Qwen tokenizer to Even's agglutinative morphology, reducing token fragmentation by 27\%.
    \item \textbf{Pipeline.} We release a complete, reproducible pipeline for adapting a vision-language model to a low-resource agglutinative language, covering tokenizer expansion, curriculum-based data synthesis, and two-stage LoRA fine-tuning.
\end{enumerate}

All datasets, evaluation benchmarks, and the complete training pipeline are released in open access. Model weights will be made available upon acceptance.

\section{Related Work}

Lupascu et al.~\cite{lupascu2026} survey multimodal models for low-resource languages and find fewer than ten papers applying VLMs to languages with under 10,000 speakers.
Prior work is concentrated on high-resource languages~\cite{qwen2025,liu2023llava,chen2023internvl}, and standard tokenizers fragment low-resource words excessively~\cite{bpe2016,rust2021tokenizer}.
Even is a ``left-behind'' language with virtually no NLP resources~\cite{joshi2020diversity}. For the purposes of this study, we rely on the primary available machine-readable sources: an Even-Russian dictionary~\cite{kan2025} and a phrasebook~\cite{sha2014}.
Recent advances in low-resource synthetic supervision, such as R2-MultiOmnia and self-training approaches for multilingual reasoning~\cite{ranaldi2025r2multiomnia, ranaldi2025multilingual}, provide promising directions for data generation. These methods leverage large teacher models to autonomously generate reasoning trajectories. While this scales well and produces rich reasoning chains, it fundamentally relies on a teacher model with acceptable performance in the target language. For severely under-resourced languages like Even, where base VLMs yield 0\% accuracy, such self-training pipelines face an insurmountable cold-start problem. In contrast, our template-based dictionary synthesis is deterministic and highly controllable. It allows us to enforce strict linguistic constraints and morphological accuracy without requiring a pre-trained teacher model, making it a more viable approach for languages at the zero-shot extreme of the resource spectrum, albeit at the cost of lower scalability compared to autonomous reasoning generation.

\section{The Even Language and Data Collection}

\subsection{Language Profile}

Even is historically also known by the exonym \textbf{Lamut}, used especially in older ethnographic literature. The language (эвенский язык) belongs to the Northern Tungusic branch of the Tungusic language family. The dataset primarily represents the \textbf{Eastern dialect group}, which is the most widely spoken dialect cluster~\cite{malchukov1995even}. A smaller portion of the data covers the \textbf{Western dialect group}, which differs from Eastern varieties in phonology and lexicon.
It is agglutinative, with words formed by concatenating a root with multiple suffixes marking case, number, possession, and verbal categories.
The Cyrillic-based alphabet includes three graphemes not present in Russian: \evchar{ӈ} (eng), \evchar{ө} (oe), and \evchar{ӫ} (o-umlaut).
These characters are absent from standard tokenizer vocabularies, motivating our adaptation in Section~\ref{sec:tokenizer}.

\subsection{Source Data}

We collected Even language data from two sources:

\begin{itemize}[leftmargin=*,nosep]
    \item \textbf{Kanyukova \& Levin (2025) Even-Russian Dictionary.} 256 pages containing headwords, translations, example sentences, and grammatical notes.
    \item \textbf{Sharina \& Kuzmina (2014) Russian-Even Phrasebook.} 87 pages of thematic dialogues, phrases, and vocabulary.
\end{itemize}

\subsection{Data Extraction Pipeline}

PDF documents were processed with GPT-4o Mini~\cite{gpt4omini} using specialized parsing prompts designed to handle complex, multi-level dictionary entries (headword $\to$ translation $\to$ example $\to$ grammatical metadata).
The structured output was manually verified and cleaned, producing a final corpus of 50,521 entries. To ensure the linguistic accuracy of our morphological generation templates and to resolve specific translation ambiguities, we consulted a linguistics specialist at a university in Yakutia who is a native Even speaker.
Each entry contains four fields: \texttt{even}, \texttt{russian}, \texttt{category}, and \texttt{source}.

\section{Data Synthesis}

\subsection{Template-Based Generation}

\begin{figure}[h!]
\centering
\includegraphics[width=0.65\columnwidth]{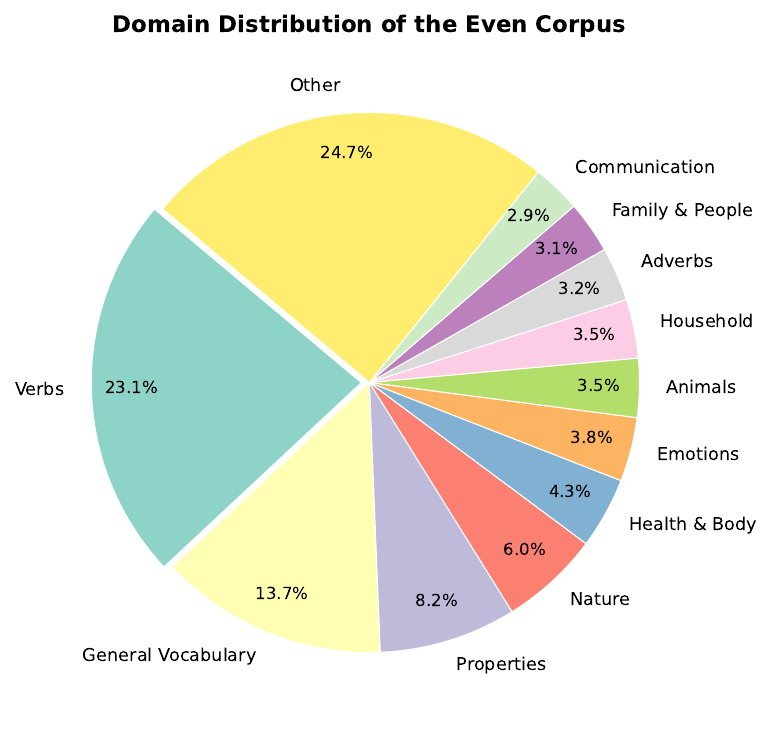}
\caption{Domain distribution of the Even corpus (50,521 entries, 31 categories).}
\label{fig:domains}
\end{figure}

From the 50,521-entry cleaned corpus, we generate training examples via templated transformations.
Each dictionary entry is expanded into multiple examples of different task types.
The full dataset comprises 107,543 training and 5,658 validation examples, organized in a three-stage curriculum:

\begin{itemize}[leftmargin=*,nosep]
    \item \textbf{Stage 1 -- Vocabulary (47,366 examples).} Word-level translation in both directions (Even$\to$Russian, Russian$\to$Even) and phrase translation.
    \item \textbf{Stage 2 -- Grammar (18,977 examples).} Declension, negation, plural formation, and verb conjugation exercises derived from morphological templates.
    \item \textbf{Stage 3 -- Comprehension (41,200 examples).} Sentence translation, fill-in-the-blank, reading comprehension, and semantic classification, including few-shot demonstrations.
\end{itemize}
The task distribution is provided in Table~\ref{tab:tasks} (Appendix~\ref{sec:appendix_details}).
Figure~\ref{fig:domains} shows the domain distribution of the source corpus, spanning 31 categories with the largest shares in verbs (23.1\%) and general vocabulary (13.7\%).

\subsection{Multimodal VQA Dataset}

We collected 189 images related to Even culture and daily life (nature, reindeer herding, hunting, clothing, tools, food, household, folklore).
Each image received bilingual (Even/Russian) question-answer annotations, yielding 477 annotation pairs.
By generating both short and long answer variants, we expanded these to 1,908 training examples.
A held-out test set of 20 examples balanced across 10 categories provides the VQA evaluation benchmark.

\section{Tokenizer Adaptation}
\label{sec:tokenizer}

The original \qwen tokenizer (151,669 tokens) fragments Even words into an average of 4.99 tokens (median 5), with only 7.6\% occupying two or fewer tokens.
The three special Cyrillic characters (\evchar{ӈ}, \evchar{ө}, \evchar{ӫ}) are absent from the vocabulary, causing further fragmentation.

\subsection{Expansion Algorithm}

We propose the following algorithm:

\paragraph{Step 1: Special graphemes.}
Add \evchar{ӈ}, \evchar{ө}, and \evchar{ӫ} as individual tokens (3 tokens).

\paragraph{Step 2: Whitelist + BPE merging.}
\begin{itemize}[nosep]
    \item \textbf{Whitelist}: 209 high-frequency Even words from the dictionary; 197 required merging (the remaining 12 were already single tokens).
    \item \textbf{VQA answers}: 101 recurrent answer words added as whole tokens.
    \item \textbf{BPE}: Iterative merging of frequent adjacent token pairs on the full Even text corpus (51,477 texts). Maximum 300 tokens total across this step, minimum pair frequency 10, over 5 rounds.
\end{itemize}

\paragraph{Step 3: Special character combinations.}
Extract bigrams and trigrams containing \evchar{ӈ}, \evchar{ө}, or \evchar{ӫ} from dictionary entries and add those meeting frequency thresholds.
In the current corpus, no combination exceeded the threshold, so this step contributed 0 tokens at this stage but remains in the pipeline for future dataset expansion.

\subsection{Results}

The expansion added 303 tokens, growing the vocabulary to 151,972.
Table~\ref{tab:tokenizer} (Appendix~\ref{sec:appendix_details}) summarizes the improvements.

The most frequent BPE tokens reflect Even's agglutinative morphology: suffix ``-н'' (5,010 occurrences), roots ``дэ'' (4,882), ``эй'' (4,140), and ``эл'' (3,338).
After expansion, \texttt{resize\_token\_embeddings(151972)} enlarges the input embeddings and language model head, initializing new rows via the default Transformers approach.

\section{Model Architecture and Training}
\label{sec:training}

\subsection{Base Model}

\qwen~\cite{qwen2025} is a 4.4B-parameter vision-language model with a Transformer decoder (text) and a ViT image encoder (336 $\times$ 336 input, 14 $\times$ 14 patches).
It supports multilingual inputs and multimodal generation natively.

\subsection{Stage 1: Text Fine-Tuning with LoRA}

\paragraph{LoRA configuration.}
We apply LoRA~\cite{lora2021} to the query, key, value, and output projection matrices of all attention layers with rank $r=32$, $\alpha=64$, and dropout 0.1.
This adds 23,592,960 trainable parameters (0.53\% of 4,461,408,768 total).
Other parameters remain frozen.
The rank is chosen as a trade-off between adaptation capacity and GPU memory (V100 32GB): $r=32$ fits within the 32 GB budget while providing sufficient capacity for text-domain adaptation.

\paragraph{Training setup.}
Training uses LoRA rank $r\!=\!32$, $\alpha\!=\!64$, dropout 0.1, effective batch size 32 (gradient accumulation $2\times$), learning rate $2\!\times\!10^{-5}$ with cosine decay, and 6 epochs on a single V100-SXM2-32GB (full details in Table~\ref{tab:stage1_hyper}, Appendix~\ref{sec:appendix_details}).

\begin{figure}[htp]
\centering
\includegraphics[width=0.85\columnwidth]{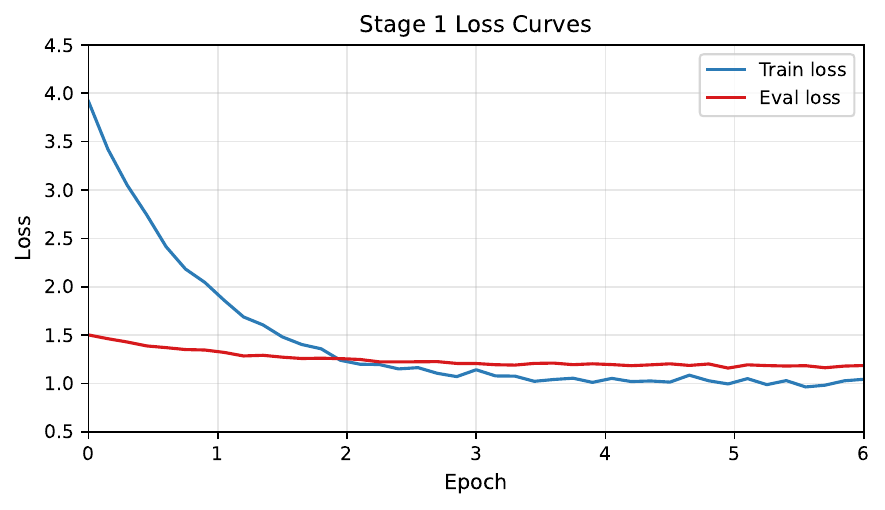}
\caption{Training and validation loss curves for Stage 1 (text LoRA fine-tuning). Validation loss reaches minimum of 1.179 at epoch 6.}
\label{fig:loss}
\end{figure}

\paragraph{Training dynamics.}
The initial loss of 3.9 decreases rapidly during the first epoch.
Final training loss reaches approximately 1.019; validation loss improves from 1.5 to a best of 1.179 at step 20,000 (Figure~\ref{fig:loss}).
Total training time is 6.8 hours.
The best checkpoint is saved, then LoRA weights are merged into the base model.

\subsection{Stage 2: VQA Fine-Tuning}

\paragraph{LoRA configuration.}
The two-stage design allows different LoRA configurations per stage: Stage 2 applies a higher rank ($r=64$, $\alpha=128$, dropout 0.05) across seven projection matrices (q, k, v, o, gate, up, down), compared to Stage 1's $r=32$ across four matrices.
This increase reflects three factors: (i) multimodal learning requires cross-modal alignment capacity beyond text-only adaptation, (ii) the frozen vision encoder places additional burden on language-side parameters to learn visual grounding, and (iii) technical constraints (GPU memory, training time) set a lower ceiling for Stage 1, where the larger text dataset (50,521 entries vs.\ 1,908 VQA pairs) would have made $r=64$ prohibitively expensive.
The vision encoder is frozen (zero trainable parameters), and we wrap its forward pass with \texttt{torch.no\_grad()} to avoid unnecessary autograd graph construction and activation storage.
Additionally, we set \texttt{vision\_model.eval()} to disable dropout in ViT.
These optimizations accelerate the forward pass by an estimated 2--3$\times$, as ViT forward was the primary bottleneck (97\% of step time prior to optimization).

\paragraph{Training setup.}
VQA training uses batch size 8 (effective 8, no gradient accumulation) for 3 epochs, learning rate $1\times10^{-4}$ with cosine decay, maximum sequence length 2,048 tokens, and image resolution capped at 512 $\times$ 512 pixels (262,144 pixels maximum).

\paragraph{Training dynamics.}
Loss decreases from 10.2 to 6.1, indicating incomplete convergence due to limited VQA data volume.
Nevertheless, the model acquires basic multimodal capability, as shown in Section~\ref{sec:results}.

\section{Evaluation and Results}
\label{sec:results}

\subsection{Benchmark Suite and Metrics}

Our text evaluation set comprises 100 examples spanning reading comprehension (RC, 30), classification (CL, 20), grammar (GR, 20), and sentence translation (TR, 30).
Standard metrics for each task type are unsuitable for an agglutinative low-resource language like Even.
BLEU~\cite{bleu2002} requires multiple reference translations and performs poorly on short sequences.
ROUGE~\cite{rouge2004} is designed for summarization, not morphological accuracy.
CER and WER operate at the character and word level but conflate root and affix errors equally.
For Even, a correct root with a wrong suffix should receive \textbf{partial credit}---this is essential for an agglutinative language where a single word encodes multiple morphological categories and full vocabulary coverage is infeasible.

For short-answer tasks (RC, CL, GR), we therefore report \textbf{Exact Match} (binary correctness, with multi-reference support via ``;'') and \textbf{Edit Similarity} (normalized Levenshtein ratio, $1 - \text{edit\_dist} / (|s_1| + |s_2|)$), classified into three categories: exact ($\geq 1.0$), partial ($\geq 0.75$), and wrong ($< 0.75$), with weights 1.0, 0.5, and 0.0 respectively. The 0.75 threshold is chosen empirically: for agglutinative Even, a correct root with one incorrect suffix yields a Levenshtein ratio of approximately 0.75--0.85, which this boundary captures as partial credit.
The \textbf{Weighted Score} is the sum of these weights across all examples.
For TR, we use \textbf{chrF} ($\beta=2$, n-grams 1--6)~\cite{chrf2015}, which operates on character n-grams and is robust to Even's relatively flexible word order while capturing morphological overlap that word-level metrics would miss.
VQA accuracy is measured via Exact Match on 20 test examples, appropriate given the short-answer format (1--3 words).

We evaluate three configurations:
\begin{enumerate}[nosep]
    \item \textbf{Base:} Untrained \qwen.
    \item \textbf{Base + Tokenizer:} Tokenizer expanded but no fine-tuning.
    \item \textbf{Base + Tokenizer + LoRA:} Full pipeline (our method).
\end{enumerate}
The base model cannot produce Even output; tokenizer expansion alone slightly degrades performance (the 303 new tokens receive randomly initialized embeddings that carry no linguistic information, adding noise during decoding); our full method substantially improves all metrics by adapting these embeddings through fine-tuning.

\begin{table}[htp]
\centering
\small
\caption{Main comparative results.
RC = Reading Comprehension, CL = Classification, GR = Grammar, TR = Sentence Translation.}
\label{tab:main_results}
\resizebox{\columnwidth}{!}{%
\begin{tabular}{lccc}
\toprule
\textbf{Metric} & \textbf{Base} & \textbf{+ Tokenizer} & \textbf{+ LoRA} \\
\midrule
Exact Match & 0.0 & 0.0 & 11.4 \\
Weighted Score & 2.1 & 0.7 & 13.6 \\
chrF & 0.0723 & 0.0694 & 0.1283 \\
\bottomrule
\end{tabular}
}
\end{table}

\FloatBarrier
\subsection{VQA Results}

Table~\ref{tab:vqa_results} reports VQA accuracy across configurations.
Text-only training (Stage 1) yields zero VQA capability---multimodal performance requires multimodal training.
Stage 2 (VQA LoRA) achieves 65\% accuracy (13/20), confirming that 1,908 examples suffice for basic visual-lexical mapping despite incomplete convergence. Note that due to the small test set size ($n=20$), these accuracy estimates have wide confidence intervals (approx. 43.2\%--82.0\% at the 95\% confidence level). Expanding these evaluation sets is a primary goal for future work.

\begin{table}[htp]
\centering
\small
\caption{VQA evaluation results (Exact Match on 20 tests).}
\label{tab:vqa_results}
\resizebox{\columnwidth}{!}{%
\begin{tabular}{lcc}
\toprule
\textbf{Configuration} & \textbf{Correct} & \textbf{Accuracy} \\
\midrule
Base & 0/20 & 0\% \\
Base + Stage 1 & 0/20 & 0\% \\
Full pipeline (Stage 1 + Stage 2) & 13/20 & 65\% \\
\bottomrule
\end{tabular}
}
\end{table}

\subsection{Qualitative Observations}

Table~\ref{tab:examples} shows sample model outputs across task types, illustrating successful predictions as well as typical errors.

\begin{table}[htp]
\centering
\small
\caption{Qualitative examples of model outputs.}
\label{tab:examples}
\setlength{\tabcolsep}{4pt}%
\renewcommand{\arraystretch}{1.4}%
\resizebox{\columnwidth}{!}{%
\begin{tabular}{>{\raggedright\arraybackslash}p{1.5em}>{\raggedright\arraybackslash}p{8em}>{\raggedright\arraybackslash}p{5em}>{\raggedright\arraybackslash}p{5em}}
\toprule
\textbf{Task} & \textbf{Input} & \textbf{Target} & \textbf{Output} \\
\midrule
CL & Оран делгэнкэ-гу? & делгэнкэ & делгэнкэ (EM) \\
VQA & Фотола як ичун? & талак & талак (EM) \\
RC & \evchar{Ӈ}и тогду тэгэттэн? & бэй & тэгэттэн (wrong) \\
TR & Я вижу медведя & Би накату иттэм & Би сэм иттэм (partial) \\
VQA & Фотола адыккан бисни? & дыгэн (4) & илан (3) (wrong) \\
\bottomrule
\end{tabular}
}
\end{table}

\paragraph{Translation.}
The model translates frequent Even words (reindeer herding, household items, nature) confidently; rare words are weaker.
Notably, the model sometimes translates words \textbf{not present in the training set}, constructing plausible Even forms from known morphological elements and analogies---a valuable property for an agglutinative language where full vocabulary coverage is infeasible.

\paragraph{Classification.}
The model performs four semantic classification types correctly: noun type (animal / plant / object), kinship (relative / non-relative), part of speech (pronoun / noun), and domain (natural / household).

\paragraph{VQA by category.}
Per-category VQA accuracy is expected to be highest for simple ``what'' questions on concrete categories (food, instruments, folklore), and lower for context-dependent questions (``what season?''), counting (``how many?''), and verbose descriptions.
The partial convergence (loss 10.2$\to$6.1) suggests substantial room for improvement with more data.

\section{Error Analysis}

We observe three error types: (1) \textbf{Semantic substitution}---the model outputs a related but incorrect word (e.g., a synonym from another dialect), indicating partial lexical acquisition; (2) \textbf{Grammatical inaccuracies}---correct roots with incorrect inflections, partially captured by Weighted Levenshtein; and (3) \textbf{Translation incompleteness} for long sentences with unfamiliar words. In VQA, the model produces semantically plausible but inexact answers (e.g., ``илан'' (three) instead of ``дыгэн'' (four)), suggesting basic visual-lexical associations that lack precision due to limited data.

\section{Conclusion and Future Work}

We presented EvenVLM, the first text and multimodal dataset for Even, with a full pipeline for fine-tuning a VLM using tokenizer adaptation and LoRA.
Using 50K dictionary entries expanded to 107K synthetic examples and 1,908 VQA examples, we adapted a 4.4B model for translation, grammar, comprehension, classification, and VQA tasks.
All datasets, benchmarks, and code are released openly.
Limitations include small VQA data volume, limited reading comprehension data, lack of parallel text corpora, Eastern dialect skew (with limited coverage of Western variants), and the need for large-scale native speaker validation of the full synthetic corpus.
Furthermore, we acknowledge that the current evaluation scale (n=20 for VQA and 100 for text benchmarks) introduces statistical noise and limits the power of our empirical claims. Expanding these evaluation sets is a primary goal for future work.
Future work includes expanding both the text corpus and the multimodal VQA dataset in size and quality, with a focus on open-access resources for model training and adaptation; involving a larger number of native Even speakers in data collection, validation, and evaluation; expanding cross-dialect representation; and extending the pipeline to Evenki---a closely related Northern Tungusic language---which shares typological features with Even and could benefit from similar adaptations.

\section*{Ethics Statement}

All dataset sources are publicly available dictionaries and phrasebooks cited in the references.
VQA images are sourced from open internet resources and cultural heritage collections.
No personal data, recordings of speakers, or sensitive information is included.
The data is released under CC-BY-NC 4.0 to protect the linguistic heritage of the Even community from commercial exploitation. We acknowledge the Even community's rights over their linguistic data and commit to engaging community representatives in future dataset iterations and validation.

\section*{Acknowledgments}

We thank the Even language community for preserving their linguistic heritage, and the creators of the Kanyukova and Levin dictionary and the Sharina and Kuzmina phrasebook.

\bibliography{references}

@article{lupascu2026,
  author = {Lupa{\c{s}}cu, M. and Rogoz, A.-C. and Stupariu, M.-S. and others},
  title   = {Large Multimodal Models for Low-Resource Languages: A Survey},
  journal = {Information Fusion},
  volume  = {131},
  pages   = {104189},
  year    = {2026},
  doi     = {10.1016/j.inffus.2026.104189},
  url     = {https://doi.org/10.1016/j.inffus.2026.104189}
}

@misc{atlas2025,
  author    = {{Interactive Atlas of Indigenous Peoples of the North, Siberia, and Far East}},
  title     = {Evens --- 2020--2021 Census Data},
  year      = {2025},
  url       = {https://atlaskmns.ru/page/en/people_eveny.html}
}

@article{qwen2025,
  author    = {Bai, S. and Cai, Y. and Chen, R. and others},
  title     = {Qwen3-VL Technical Report},
  journal   = {arXiv preprint},
  year      = {2025},
  url       = {https://arxiv.org/pdf/2511.21631}
}

@inproceedings{lora2021,
  author    = {Hu, E. J. and Shen, Y. and Wallis, P. and others},
  title     = {{LoRA}: Low-Rank Adaptation of Large Language Models},
  booktitle = {Proceedings of ICLR},
  year      = {2021},
  url       = {https://arxiv.org/abs/2106.09685}
}

@book{kan2025,
  author    = {Kanyukova, U. V. and Levin, V. I.},
  title     = {Even-Russian Dictionary (Эвенско-русский словарь)},
  publisher = {Almaz-Graf, St.\ Petersburg},
  year      = {2025},
  pages     = {256}
}

@book{sha2014,
  author    = {Sharina, S. I. and Kuzmina, R. P.},
  title     = {Russian-Even Phrasebook (Русско-эвенский разговорник)},
  publisher = {IGIiPMNS SO RAN, Yakutsk},
  year      = {2014},
  pages     = {87}
}

@inproceedings{bpe2016,
  author    = {Sennrich, R. and Haddow, B. and Birch, A.},
  title     = {Neural Machine Translation of Rare Words with Subword Units},
  booktitle = {Proceedings of the 54th Annual Meeting of the Association for Computational Linguistics (Volume 1: Long Papers)},
  year      = {2016},
  pages     = {1715--1725},
  url       = {https://aclanthology.org/P16-1162/}
}

@misc{gpt4omini,
  author    = {{OpenAI}},
  title     = {{GPT-4o Mini}: Advancing Cost-Efficient Intelligence},
  year      = {2024},
  url       = {https://openai.com/index/gpt-4o-mini-advancing-cost-efficient-intelligence}
}

@inproceedings{chrf2015,
  author    = {Popovi\'{c}, M.},
  title     = {{chrF}: character n-gram {F}-score for automatic {MT} evaluation},
  booktitle = {Proceedings of the Tenth Workshop on Statistical Machine Translation},
  year      = {2015},
  pages     = {392--395},
  url       = {https://aclanthology.org/W15-3049/}
}

@article{liu2023llava,
  author    = {Liu, H. and Li, C. and Wu, Q. and Lee, Y. J.},
  title     = {Improved Baselines with Visual Instruction Tuning},
  journal   = {Advances in Neural Information Processing Systems (NeurIPS)},
  year      = {2023},
  url       = {https://arxiv.org/abs/2310.03744}
}

@article{chen2023internvl,
  author    = {Chen, Z. and Wu, J. and Wang, W. and Su, W. and Chen, G. and others},
  title     = {{InternVL}: Scaling up Vision Foundation Models and Aligning for Generic Visual-Linguistic Tasks},
  journal   = {arXiv preprint},
  year      = {2023},
  url       = {https://arxiv.org/abs/2312.14238}
}

@inproceedings{joshi2020diversity,
  author    = {Joshi, P. and Santy, S. and Budhiraja, A. and Bali, K. and Choudhury, M.},
  title     = {The State and Fate of Linguistic Diversity and Inclusion in the {NLP} World},
  booktitle = {Proceedings of ACL},
  year      = {2020},
  url       = {https://aclanthology.org/2020.acl-main.560/}
}

@inproceedings{rust2021tokenizer,
  author    = {Rust, P. and Pfeiffer, J. and Vuli\'{c}, I. and Ruder, S. and Gurevych, I.},
  title     = {How Good is Your Tokenizer? On the Monolingual Performance of Multilingual Language Models},
  booktitle = {Proceedings of ACL},
  year      = {2021},
  url       = {https://aclanthology.org/2021.acl-long.243/}
}

@book{malchukov1995even,
  author    = {Malchukov, A. L.},
  title     = {Even},
  series    = {Languages of the World/Materials 12},
  publisher = {LINCOM Europa},
  year      = {1995},
  address   = {M\"{u}nchen}
}

@inproceedings{bleu2002,
  author    = {Papineni, K. and Roukos, S. and Ward, T. and Zhu, W.-J.},
  title     = {{BLEU}: a Method for Automatic Evaluation of Machine Translation},
  booktitle = {Proc. of ACL},
  year      = {2002},
  url       = {https://aclanthology.org/www.mt-archive.info/IBM-2001-Papineni.pdf}
}

@inproceedings{rouge2004,
  author    = {Lin, C.-Y.},
  title     = {{ROUGE}: A Package for Automatic Evaluation of Summaries},
  booktitle = {Proc. of ACL Workshop},
  year      = {2004},
  url       = {https://aclanthology.org/W04-1013/}
}

@inproceedings{ranaldi2025r2multiomnia,
  author    = {Ranaldi, L. and Ranaldi, F. and Pucci, G.},
  title     = {{R2-MultiOmnia}: Leading Multilingual Multimodal Reasoning via Self-Training},
  booktitle = {Proceedings of the 63rd Annual Meeting of the Association for Computational Linguistics (ACL)},
  year      = {2025},
  url       = {https://aclanthology.org/2025.acl-long.402/}
}

@inproceedings{ranaldi2025multilingual,
  author    = {Ranaldi, L. and Pucci, G.},
  title     = {Multilingual Reasoning via Self-Training},
  booktitle = {Proceedings of the 2025 Conference of the North American Chapter of the Association for Computational Linguistics (NAACL)},
  year      = {2025},
  url       = {https://aclanthology.org/2025.naacl-long.577/}
}

\appendix
\section{Additional Details}
\label{sec:appendix_details}

Tables~\ref{tab:tasks}, \ref{tab:tokenizer}, \ref{tab:stage1_hyper}, and \ref{tab:stage2_hyper} provide details on dataset composition, tokenizer expansion, and training configurations.

\begin{table}[h!]
\centering
\small
\caption{Training dataset composition. Total: 107,543 examples.}
\label{tab:tasks}
\resizebox{\columnwidth}{!}{%
\begin{tabular}{lrr}
\toprule
\textbf{Task} & \textbf{Count} & \textbf{Stage} \\
\midrule
Translation (word) & 44,391 & 1 \\
Translation (sentence) & 27,365 & 3 \\
Translation (phrase) & 13,363 & 1 \\
Grammar (declension) & 8,022 & 2 \\
Grammar (negation) & 4,526 & 2 \\
Grammar (plural/conjugation) & 3,876 & 2 \\
Reading comprehension & 3,000 & 3 \\
Classification & 3,000 & 3 \\
Other & 2,198 & 1--3 \\
\bottomrule
\end{tabular}%
}
\end{table}

\begin{table}[h!]
\centering
\small
\caption{Tokenizer expansion results. Fragmentation reduced by 27\%.}
\label{tab:tokenizer}
\resizebox{\columnwidth}{!}{%
\begin{tabular}{lrr}
\toprule
\textbf{Metric} & \textbf{Before} & \textbf{After} \\
\midrule
Vocabulary size & 151,669 & 151,972 \\
Avg.\ tokens/word & 4.99 & 3.64 \\
Median tokens/word & 5 & 4 \\
Words $\leq$ 2 tokens & 7.6\% & 15.4\% \\
\bottomrule
\end{tabular}%
}
\end{table}

\begin{table}[h!]
\centering
\small
\caption{Stage 1 (text) training hyperparameters.}
\label{tab:stage1_hyper}
\begin{tabular}{ll}
\toprule
\textbf{Parameter} & \textbf{Value} \\
\midrule
Effective batch size & 32 (16 $\times$ 2 grad.\ accum.) \\
Epochs & 6 \\
Learning rate & $2\times10^{-5}$ \\
LR scheduler & Cosine (3\% warmup) \\
Weight decay & 0.01 \\
Max sequence length & 512 \\
Mixed precision & fp16 \\
Gradient checkpointing & Yes \\
Save/eval frequency & Every 2,000 steps \\
\bottomrule
\end{tabular}
\end{table}

\begin{table}[h!]
\centering
\small
\caption{Stage 2 (VQA) training hyperparameters.}
\label{tab:stage2_hyper}
\begin{tabular}{ll}
\toprule
\textbf{Parameter} & \textbf{Value} \\
\midrule
LoRA rank $r$ & 64 \\
LoRA $\alpha$ & 128 \\
LoRA dropout & 0.05 \\
Target modules & q, k, v, o, gate,\\ up, down projections \\
Effective batch size & 8 \\
Epochs & 3 \\
Learning rate & $1\times10^{-4}$ \\
Vision encoder & Frozen + \texttt{torch.no\_grad()} \\
Max pixels & 262,144 ($\sim$512$\times$512) \\
Max sequence length & 2,048 \\
VQA train examples & 1,908 \\
\bottomrule
\end{tabular}
\end{table}

\end{document}